\documentclass{bmvc2k}

\usepackage[noend]{algpseudocode}
\usepackage{wrapfig}
\usepackage[linesnumbered,ruled,vlined]{algorithm2e}
\SetKwInput{KwInput}{Input}                
\SetKwInput{KwOutput}{Output} 

\title{BOAT: Bilateral Local Attention Vision Transformer}

\addauthor{Tan Yu}{tan.yu1503@gmail.com}{1}
\addauthor{Gangming Zhao}{gangmingzhao@gmail.com}{2}
\addauthor{Ping Li}{pingli98@gmail.com}{1}
\addauthor{Yizhou Yu}{yizhouy@acm.org}{2}
\addinstitution{
Cognitive Computing Lab\\ Baidu Research\\10900 NE 8th St. Bellevue, WA USA
}
\addinstitution{
Department of Computer Science\\ The University of Hong Kong
}

\runninghead{Yu et al.}{Bilateral Local Attention Vision Transformer}

\usepackage{amsfonts}
\usepackage{amsmath}
\usepackage{amsthm}
\usepackage{booktabs}
\begin{document}

\maketitle

\begin{abstract}
Vision Transformers  achieved outstanding performance in many computer vision tasks. Early Vision Transformers such as ViT and DeiT adopt global self-attention, which is computationally expensive when the number of patches is large. To improve the efficiency, recent Vision Transformers adopt local self-attention mechanisms, where self-attention is computed within local windows. Despite the fact that window-based  local self-attention significantly boosts efficiency, it fails to capture the relationships between distant but similar patches in the image plane. To overcome this limitation of image-space local attention, in this paper, we further exploit the locality of patches in the feature space. We group the patches into multiple clusters using their features, and self-attention is computed within every cluster. Such feature-space local attention effectively captures the connections between patches across different local windows but still relevant. We propose a Bilateral lOcal Attention vision Transformer (BOAT), which integrates feature-space local attention with image-space local attention. We further integrate BOAT with both Swin and CSWin models, and extensive experiments on several benchmark datasets demonstrate that our BOAT-CSWin model clearly and consistently outperforms existing state-of-the-art CNN models and vision Transformers.\vspace{-0.1in}
\end{abstract}

\section{Introduction}
\label{sec:intro}

Following the great success of Transformers~\cite{vaswani2017attention} in natural language processing tasks, researchers have recently proposed vision Transformers~\cite{dosovitskiy2021image,touvron2020training,liu2021swin,chu2021twins,chen2021mvt,yu2022degenerate}, which have achieved outstanding performance in many computer vision tasks, including image recognition, detection, and segmentation. As early versions of vision transformers, ViT~\cite{dosovitskiy2021image} and DeiT~\cite{touvron2020training} uniformly divide an image into $16\times 16$ patches (tokens) and apply a stack of standard Transformer layers to a sequence of tokens formed using these patches. The original self-attention mechanism is global, i.e., the receptive field of a patch in ViT and DeiT covers all patches of the image, which is vital for modeling long-range interactions among patches. On the other hand, the global nature of self-attention imposes a great challenge in efficiency. Specifically, the computational complexity of self-attention is quadratic in terms of the number of patches. 
As the number of patches is inversely proportional to the patch size when the size of the input image is fixed, the computational cost forces ViT and DeiT to adopt medium-size patches, which might not be as effective as smaller patches generating higher-resolution feature maps, especially for  dense prediction tasks such as segmentation.

To maintain higher resolution feature maps while achieving  high efficiency, some methods~\cite{liu2021swin,chu2021twins} exploit image-space local attention.  They divide an image into multiple  local windows, each of which includes a number of patches. Self-attention operations are only performed on patches within the same local window. This is a reasonable design since a patch is likely to be affiliated with other patches in the same local window but not highly relevant to patches in other windows. Thus, pruning attention between patches from different windows might not significantly deteriorate the performance. Meanwhile, the computational cost of window-based self-attention is much lower than that of the original self-attention over the entire image. Swin Transformer~\cite{liu2021swin} and Twins~\cite{chu2021twins} are such examples. Swin Transformer~\cite{liu2021swin} performs  self-attention within local windows. To facilitate communication between patches from different windows, Swin Transformer has two complementary window partitioning schemes, and a window in one scheme overlaps with multiple windows in the second scheme. Twins~\cite{chu2021twins} performs self-attention within local windows and builds connections among different windows by performing (global) self-attention over feature vectors sparsely sampled from the entire image using a regular subsampling pattern.

\begin{wrapfigure}{r}{0.53\textwidth}
  \centerline{\includegraphics[scale=0.8]{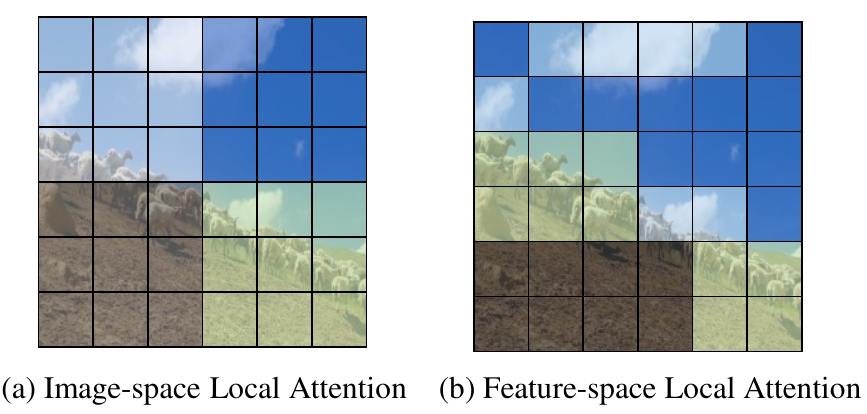}}
  
\vspace{-0.05in}

  \caption{The image-space local attention versus the feature-space local attention.}
  \label{fig:intro}\vspace{-0.1in}
\end{wrapfigure}

In this work, we rethink local attention and explore locality from a broader perspective. Specifically, we investigate feature-space local attention apart from its image-space counterpart.
Instead of computing local self-attention in the image space, feature-space local attention exploits locality in the feature space.
It is based on the fact that patch feature vectors close to each other in the feature space tend to have more influence on each other in the computed self-attention results. This is because the actual contribution of a feature vector to the self-attention result at another feature vector is controlled by the similarity between these two feature vectors. Feature-space local attention computes the self-attention result at a feature vector using its feature-space nearest neighbors only while setting the contribution from feature vectors farther away to zero. This essentially defines a piecewise similarity function, which clamps the similarity between feature vectors far apart to zero.  In comparison to the aforementioned image-space local attention, feature-space local attention has been rarely exploited in vision transformers. As shown in Figure~\ref{fig:intro}, feature-space local attention computes attention among relevant patches which might not be close to each other in the image plane. Thus, it is a natural compensation to image-space local attention, which might miss meaningful connections between patches residing in different local windows. 

In this paper, we propose a novel vision Transformer architecture, Bilateral lOcal Attention vision Transformer (BOAT), to exploit the complementarity between feature-space and image-space local attention. The essential component in our network architecture is the  bilateral local attention block, consisting of a feature-space local attention module and an image-space local attention module. The image-space local attention module divides an image into multiple local windows as Swin~\cite{liu2021swin} and CSWin~\cite{dong2022cswin}, and self-attention is computed within each local window. In contrast, feature-space local attention groups all the patches into multiple clusters and self-attention is computed within each cluster. Feature-space local attention could be implemented in a straightforward way using K-means clustering. Nevertheless, K-means clustering cannot ensure the generated clusters are evenly sized, thus impedes efficient parallel implementation. In addition, sharing self-attention parameters among unevenly sized clusters may also negatively impact the effectiveness of self-attention. To overcome this obstacle, we propose hierarchical balanced clustering, which groups patches into clusters of equal size.

We conduct experiments on multiple computer vision tasks, including image classification, semantic segmentation, and object detection. Experiments on several public benchmarks demonstrate that our BOAT clearly and consistently improves existing image-space local attention vision Transformers, including Swin~\cite{liu2021swin} and CSWin~\cite{dong2022cswin}, on these tasks.

\vspace{-0.15in}
\section{Related Work}\label{sec:related}

\vspace{-0.1in}
\subsection{Vision Transformers}
\vspace{-0.1in}

In the past decade, CNN has achieved tremendous successes in  numerous computer vision tasks~\cite{krizhevsky2012imagenet,he2016deep}.
The  natural language processing (NLP) backbone, Transformer, has recently attracted the attention of researchers in the computer vision community. After dividing an image into non-overlapping patches (tokens), Vision Transformer (ViT)~\cite{dosovitskiy2021image} applies  Transformer for communications among the tokens.  Without delicately devised convolution kernels, ViT achieved excellent performance in image recognition in comparison to  CNNs using a huge training corpus.
DeiT~\cite{touvron2020training}  improves data efficiency by exploring  advanced training and data augmentation strategies. Recently, many efforts have been devoted to  improving the recognition accuracy and efficiency of Vision Transformers.

To boost the recognition accuracy,  T2T-ViT~\cite{yuan2021tokens} proposes a  Tokens-to-Token transformation, recursively aggregating neighboring tokens into one token for  modeling local structures.
 TNT~\cite{han2021transformer} also investigates local structure modeling. It additionally builds an inner-level Transformer to model the visual content within each local patch.
PVT~\cite{wang2021pyramid} uses small-scale patches, yielding higher resolution feature maps for dense prediction. Meanwhile, PVT progressively shrinks the feature map size for computation reduction. PiT~\cite{heo2021rethinking} also decreases spatial dimensions through pooling and  increases channel dimensions in deeper layers. 

More recently,  computing self-attention within local windows~\cite{liu2021swin,chu2021twins,huang2021shuffle,dong2022cswin}, has achieved a good trade-off between effectiveness and efficiency. For example, Swin~\cite{liu2021swin} divides an image into multiple  local windows and  computes self-attention among patches from the same window.  To achieve communication across local windows, Swin  shifts window configurations in different layers. Twins~\cite{chu2021twins} also exploits local windows for enhancing efficiency. To achieve cross-window communication, it computes additional self-attention over features sampled from the entire image. Similarly, Shuffle Transformer~\cite{huang2021shuffle} exploits local windows and performs cross-window communication by shuffling patches.
CSWin~\cite{dong2022cswin} adopts cross-shaped windows,  computing self-attention in horizontal and vertical stripes in parallel. The aforementioned local attention models~\cite{liu2021swin,chu2021twins,huang2021shuffle,dong2022cswin} only exploit image-space locality. In contrast, our BOAT exploits not only image-space locality but also feature-space locality.

\subsection{Efficient Transformers}

High computational costs limit Transformer's usefulness in practice. Thus, much research~\cite{tay2020efficient} has recently been dedicated to improving efficiency.  One popularly used strategy for speeding up Transformers enforces sparse attention matrices by limiting the receptive field of each token. Image Transformer~\cite{parmar2018image} and Block-wise Transformer~\cite{qiu2020blockwise} divide a long sequence into local buckets. In this case, the attention matrix has a block-diagonal structure. Only self-attention within each bucket is retained, and cross-bucket attention is pruned.
Transformers based on image-space local attention, such as Swin Transformer~\cite{liu2021swin}, Twins~\cite{chu2021twins}, Shuffle Transformer~\cite{huang2021shuffle}, and CSWin Transformer~\cite{dong2022cswin} also adopt buckets (windows) for boosting efficiency. In parallel to bucket-based local attention, strided attention is another approach for achieving sparse attention matrices. Sparse Transformer~\cite{child2019generating} and LongFormer~\cite{beltagy2020longformer} utilize strided attention, which computes self-attention over features sampled with a sparse grid with a stride larger than one, leading to a sparse attention matrix facilitating faster computation. The global sub-sampling layer in Twins~\cite{chu2021twins} and the shuffle module in Shuffle Transformer~\cite{huang2021shuffle} can be regarded as strided attention modules. Some recent works exploit pure MLP-based architectures~\cite{yu2021rethinking,yu2022s2mlp,yu2022s2mlpv2,chen2022r2mlp} to boost efficiency.

Unlike the above mentioned image-space local attention, several methods determine the scope of local attention in the feature space. Reformer~\cite{kitaev2020reformer} distributes tokens to buckets by feature-space hashing functions. Routing Transformer~\cite{roy2021efficient} applies online K-means to cluster tokens. Sinkhorn Sorting Network~\cite{tay2020sparse} learns to sort and divide an input sequence into  chunks. Our feature-space local attention module also falls into this category. As far as we know, this paper is the first attempt to apply feature-space grouping to vision Transformers.

\section{Method}
\label{sec:met}


\begin{figure*}[h]
    \centering
    \includegraphics[scale=0.55]{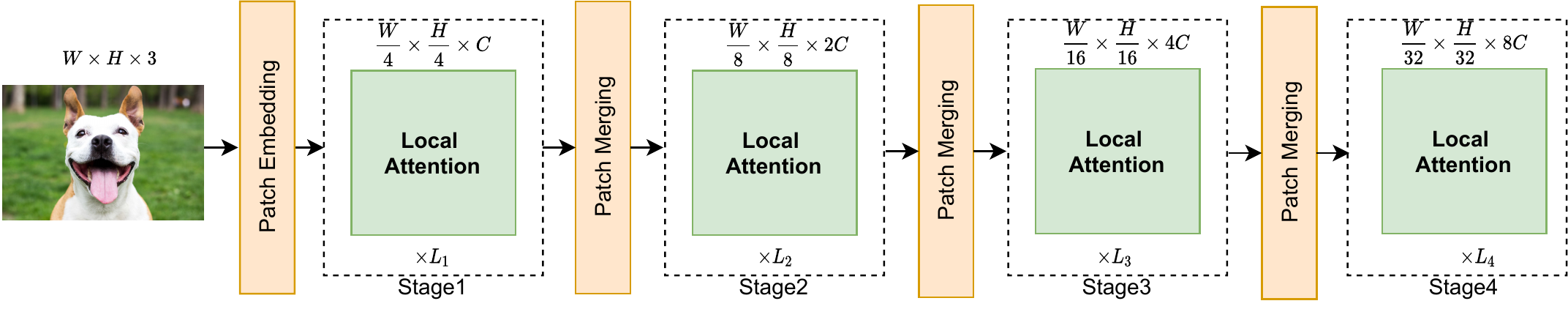}
     \vspace{-0.05in}
    \caption{Architecture of  Bilateral lOcal Attention Vision Transformer (BOAT). }
    \label{fig:arc}
    \vspace{-0.1in}
\end{figure*}

As visualized in Figure~\ref{fig:arc}, the  proposed BOAT architecture consists of a patch embedding module, and a stack of $L$ Bilateral Local Attention blocks. Meanwhile, we exploit a hierarchical pyramid structure. Below we only briefly introduce the patch embedding module and the hierarchical pyramid structure and leave the details of  the proposed Bilateral Local Attention block in Section~\ref{sec:clab} and \ref{sec:cla}.


\vspace{1mm}
\noindent \textbf{Patch embedding.}  For an input image with size $H \times W$, we follow Swin~\cite{liu2021swin} and CSWin Transformer~\cite{dong2022cswin}, and leverage convolutional token embedding ($7 \times 7$ convolution layer with stride 4) to obtain $\frac{H}{4} \times \frac{W}{4}$ patch tokens, and the dimension of each token is $C$.

\vspace{1mm}
\noindent \textbf{Hierarchical pyramid structure.} Similar to Swin~\cite{liu2021swin} and CSWin Transformer~\cite{dong2022cswin}, we also build a hierarchical pyramid structure. The whole architecture consists of four stages. A convolution layer ($3 \times 3$, stride $2$) is used between two adjacent stages to merge patches. It reduces the number of tokens and doubles the number of channels. Therefore, in the $i$-th stage, the feature map contains $\frac{H}{2^{(i+1)}} \times \frac{W}{2^{(i+1)}}$ tokens and  $2^{i-1}C$ channels.

\subsection{Bilateral Local Attention Block} 
\label{sec:clab}

\begin{figure}[h]

\vspace{-0.1in}

    \centering
    \includegraphics[scale=0.7]{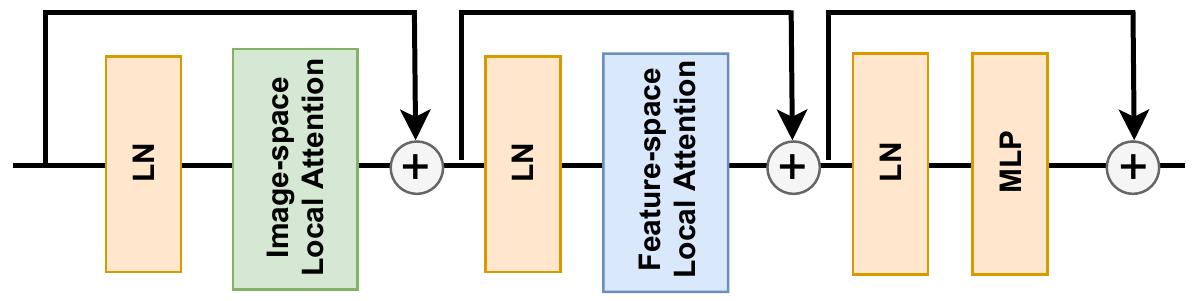}
    \vspace{-0.1in}
    \caption{Architecture of Bilateral Local Attention (BLA) Block.}
    \label{fig:str}\vspace{-0.1in}
\end{figure}

As shown in Figure~\ref{fig:str},  a Bilateral Local Attention (BLA) Block consists of an image-space local attention (ISLA) module, a feature-space (content-based) local attention (FSLA) module, an MLP module, and several layer normalization (LN) modules.
Let us denote the set of input tokens by $\mathcal{T}_{\mathrm{in}} = \{\mathbf{t}_i\}_{i=1}^N$ where  $\mathbf{t}_i \in \mathbb{R}^{C}$, $C$ is the number of channels and $N$ is the number of tokens.
The input tokens go  through a normalization layer followed by an image-space local attention (ISLA) module, which has a shortcut connection:
\begin{equation}
    {\mathcal{T}}_{\mathrm{ISLA}} = \mathcal{T}_{\mathrm{in}} + \mathrm{ISLA}(\mathrm{LN}(\mathcal{T}_{\mathrm{in}})).
\end{equation}
Image-space local attention only computes self-attention among tokens within the same local window. We adopt existing window-based local attention modules, such as those in Swin Transformer~\cite{liu2021swin} and CSWin Transformer~\cite{dong2022cswin} as our ISLA module due to their excellent performance. 
Intuitively, patches within the same local window are likely to be closely related to each other. However, some distant patches in the image space might also reveal important connections, such as similar contents, that could be helpful for visual understanding. Simply throwing away such connections between distant patches in the image space might deteriorate image recognition performance.

To bring back the useful information dropped out by image-space local attention, we develop a feature-space local attention (FSLA) module. 

The output of the ISLA module, ${\mathcal{T}}_{\mathrm{ISLA}}$, is fed into another normalization layer followed by a feature-space (content-based) local attention (FSLA) module, which also has a shortcut connection:
\begin{equation}
    {\mathcal{T}}_{\mathrm{FSLA}} = {\mathcal{T}}_{\mathrm{ISLA}} + \mathrm{FSLA}(\mathrm{LN}({\mathcal{T}}_{\mathrm{ISLA}})).
\end{equation}
The FSLA module computes self-attention among tokens that are close in the feature space, which is  complementary to the ISLA module. Meanwhile, by only considering local attention in the feature space, FSLA is more efficient than the original (global) self-attention.  We will present the details of FSLA in Section~\ref{sec:cla}.  Following CSWin~\cite{dong2022cswin}, we also add locally-enhanced positional encoding to each feature-space local attention layer to model position.

At last, the output of the FSLA module,  ${\mathcal{T}}_{\mathrm{FSLA}}$, is processed by another normalization layer and an MLP module to generate the output of a Bilateral Local Attention Block:
\begin{equation}
    {\mathcal{T}}_{\mathrm{out}} = {\mathcal{T}}_{\mathrm{FSLA}}  + \mathrm{MLP}(\mathrm{LN}({\mathcal{T}}_{\mathrm{FSLA}})).
\end{equation}
Following existing vision Transformers~\cite{dosovitskiy2021image,touvron2020training}, the MLP module consists of two fully-connected layers. The first one increases the feature dimension from $C$ to $rC$ and the second one decreases the feature dimension from $rC$ back to $C$. By default, we set $r=4$.

\vspace{-0.05in}
\subsection{Feature-Space Local Attention}
\label{sec:cla}
Different from image-space local attention which groups tokens according to their spatial locations in the image plane, feature-space local attention seeks to group tokens according to their content, \emph{i.e.}, features.
We could simply perform K-means clustering on token features to achieve this goal. Nevertheless, K-means clustering cannot ensure that the generated clusters are equally sized, which makes it difficult to have efficient parallel implementation on GPU platforms, and may also negatively impact the overall effectiveness of self-attention. 

\vspace{0.1in}

\begin{wrapfigure}{r}{0.44\textwidth}\vspace{-0.2in}
  \centerline{\includegraphics[width=2.1in]{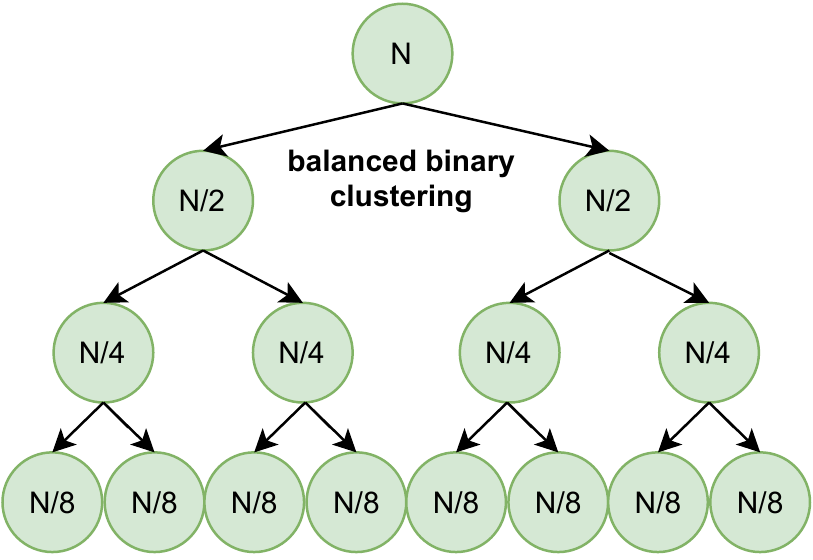}}
  \caption{Example of balanced hierarchical clustering. In this example, the number of hierarchical levels is $3$. There are $2^3 =8$ clusters in the bottom level.\vspace{-0mm}}
  \label{fig:hc}\vspace{-0.1in}
\end{wrapfigure}

\noindent\textbf{Balanced hierarchical clustering.} To overcome the imbalance problem of K-means clustering, we propose a balanced hierarchical clustering, which performs $K$ levels of clustering. At each level, it conducts balanced binary clustering, which equally splits a set of tokens into two clusters. Let us denote the set of input tokens by ${\mathcal{T}} = \{\mathbf{t}_i\}_{i=1}^N$.  In the first level, it splits $N$ tokens in $\mathcal{T}$ into two subsets with ${N}/{2}$ tokens each. At the $k$-th level, it splits  $N/2^{k-1} $ tokens assigned to the same subset in the upper level into two smaller subsets of $N/2^k$ size. At the end, we obtain $2^K$ evenly sized subsets in the final level, $\{\mathcal{T}_i\}_{i=1}^{2^K}$, and the size of each subset $|\mathcal{T}_i|$ is equal to $N/{2^K}$. Here, we require the condition that $N$ is divisible by $2^K$, which can be easily satisfied in existing vision Transformers.
We visualize the process of balanced hierarchical clustering in Figure~\ref{fig:hc}.
The core operation in balanced hierarchical clustering is our devised balanced binary clustering, which we elaborate below.

\vspace{0.05in}
\noindent\textbf{Balanced binary clustering.}
Given a set of $2m$ tokens $\{\mathbf{t}_i\}_{i=1}^{2m}$, balanced binary clustering divides them into two groups and the size of each group is $m$. 
Similar to K-means clustering, our balanced binary clustering relies on cluster centroids. To determine the cluster membership of each sample, K-means clustering only considers the distance between the sample and all centroids.
In contrast, our balanced binary clustering further requires that the two resulting clusters have equal size. Let us denote the two cluster centroids as $\mathbf{c}_1$ and $\mathbf{c}_2$. For each token $\mathbf{t}_i$, we compute distance ratio, $r_i$, as a metric to determine its cluster membership:
\begin{equation}
    r_i = \frac{s(\mathbf{t}_i, \mathbf{c}_1)}{ s(\mathbf{t}_i, \mathbf{c}_2)},~\forall i\in[1,2m],
\end{equation}
where $s(\mathbf{x},\mathbf{y})$ denotes the cosine similarity between $\mathbf{x}$ and $\mathbf{y}$. The $2m$ tokens $\{\mathbf{t}_i\}_{i=1}^{2m}$ are sorted in the decreasing order of their distance ratios $\{{r}_i\}_{i=1}^{2m}$. We assign the tokens in the first half of the sorted list to the first cluster $\mathcal{C}_1$ and those in the second half of the sorted list to the second cluster $\mathcal{C}_2$, where the size of both $\mathcal{C}_1$ and $\mathcal{C}_2$ is $m$. The mean of the tokens from each cluster is used to update the cluster centroid. Similar to K-means, our balanced binary clustering updates cluster centroids and the cluster membership of every sample in an iterative manner. Note that cluster centroids are always computed on the fly, and are not learnable parameters. The detailed steps of the proposed balanced binary clustering are given in Algorithm~\ref{alg:1}.

\begin{algorithm}[h]
\DontPrintSemicolon
  
  \KwInput{Tokens $\{\mathbf{t}_i\}_{i=1}^{2m}$ and the iteration number,  $T$. }
  \KwOutput{Two clusters, $\mathcal{C}_1$ and $\mathcal{C}_2$.}
 
   Initialize centroids $\mathbf{c}_1 = \frac{\sum_{i=1}^m \mathbf{t}_i}{m}$, $\mathbf{c}_2 = \frac{\sum_{i=m+1}^{2m} \mathbf{t}_i}{m}$

   \While{$n\_iter \in [1,T]$}
   {
   		\For{$i \in [1,2m]$} 
   		  {$r_i = \frac{s(\mathbf{t}_i, \mathbf{c}_1)}{ s(\mathbf{t}_i, \mathbf{c}_2)}$
   		  }
   		[$i_1,\cdots,i_{2m}$] =  argsort([$r_1,\cdots, r_{2m}$])
   		
   		$\mathcal{C}_1 = \{\mathbf{t}_{i_j}\}_{j=1}^m$, $\mathcal{C}_2 = \{\mathbf{t}_{i_j}\}_{j=m+1}^{2m}$
   		
   		$\mathbf{c}_1 = \frac{\sum_{\mathbf{c} \in \mathcal{C}_1} \mathbf{c}}{m}$,  $\mathbf{c}_2 = \frac{\sum_{\mathbf{c} \in \mathcal{C}_2} \mathbf{c}}{m} $
   }

\caption{Balanced Binary Clustering.}
\label{alg:1}
\end{algorithm}

In the aforementioned  balanced binary clustering,  two resulting clusters have no shared tokens, \emph{i.e.}, $\mathcal{C}_1 \cap \mathcal{C}_2 = \emptyset$. One main drawback of the non-overlapping setting is that, a token in the middle portion of the sorted list has some of its feature-space neighbors in one cluster while the other neighbors in the other cluster. No matter which cluster this token is finally assigned to, the connection between the token and part of its feature-space neighbors will be cut off. For example, the token at the $m$-th location of the sorted list cannot communicate with the token at the $m+1$-st location during attention calculation because they are assigned to different clusters.
Overlapping balanced binary clustering overcomes this drawback by assigning the first $m+n$ tokens in the sorted list to the first cluster, \emph{i.e.}, $\hat{\mathcal{C}}_1 = \{\mathbf{t}_{j_i}\}_{i=1}^{m+n}$, and the last $m+n$ tokens in the sorted list to the second cluster, \emph{i.e.}, $\hat{\mathcal{C}}_2 = \{\mathbf{t}_{j_i}\}_{i=m-n+1}^{2m}$. Thus, the two resulting clusters have $2n$ tokens in common, \emph{i.e.}, $\hat{\mathcal{C}}_1 \cap \hat{\mathcal{C}}_2 = \{\mathbf{t}_{j_i}\}_{i=m-n+1}^{m+n}$. 
By default,  we only adopt overlapping binary clustering at the last level of the proposed balanced hierarchical clustering and use the non-overlapping version at the other levels. We set  $n=20$ in all experiments for overlapping binary clustering.

\vspace{0.05in}
\noindent\textbf{Local attention within cluster.}
Through the above introduced balanced hierarchical clustering, the set of tokens, $\mathcal{T}$, are grouped into $2^K$ subsets $\{\mathcal{T}_{i}\}_{i=1}^{2^K}$, where $|\mathcal{T}_{i}| = \frac{N}{2^K}$.

The standard self-attention (SA) is performed within each subset:
\begin{equation}
\label{eq:mhsa}
   \hat{\mathcal{T}}_k  = \mathrm{SA}({\mathcal{T}}_k),~\forall k\in[1,2^K].
\end{equation}
The output, $\hat{\mathcal{T}}$, is the union of all attended subsets:
\begin{equation}
 \hat{\mathcal{T}} =  \underset{k\in[1,K]}{\bigcup} \hat{\mathcal{T}}_k.  
\end{equation}
Following the multi-head configuration in Transformer, we also devise multi-head feature-space local attention. Note that, in our multi-head feature-space local attention, we implement multiple heads not only for computing self-attention in Eq.~(\ref{eq:mhsa}) as a standard Transformer, but also for performing balanced hierarchical clustering. That is, balanced hierarchical clustering is performed independently in each head. Thus, for a specific token, in different heads, it might pay feature-based local attention to different tokens. This configuration is more flexible than Swin~\cite{liu2021swin}, where multiple heads share the same local window.

\vspace{-0.05in}
\section{Experiments}\label{sec:exp}
\vspace{-0.05in}

To demonstrate the effectiveness of our BOAT  as a general vision backbone, we conduct experiments on image classification, semantic segmentation and object detection.

We build  BOAT on top of two recent local attention vision Transformers, Swin~\cite{liu2021swin} and CSWin~\cite{dong2022cswin}. We term the BOAT built upon Swin as BOAT-Swin. 
In BOAT-Swin, the image-space local attention (ISLA) module adopts shifted  window  attention in Swin. In contrast, the ISLA module in BOAT-CSWin uses cross-shape window attention in CSWin. We provide the detailed specifications of BOAT-Swin and BOAT-CSWin in Section 1 of the supplementary materials. Meanwhile, we present main experimental results in the following sections. More ablation studies  are presented in Section 2 of the supplementary materials. 

\vspace{-0.05in}
\subsection{Image Classification} 
\vspace{-0.05in}

\begin{table}[h]\small
	\setlength{\abovecaptionskip}{0pt}%
	\setlength{\belowcaptionskip}{0pt}
\begin{center}
\resizebox{0.98\linewidth}{!}
{
\begin{tabular}{@{}lcccc|lcccc@{}}
\toprule[1pt]
Method  & size& \#para.  & FLOPs  & Top-1 &  Method  & size& \#para.  & FLOPs  & Top-1 \\ 
\midrule
ReGNetY-4G~\cite{radosavovic2020designing} & 224 & 21M & 4.0G & 80.0 &
 & & & & \\ 
PVTv2-B2~\cite{wang2021pvtv2} & 224 & 25M & 4.0G & 82.0 &
Focal-T~\cite{yang2021focal} & 224 &  29M & 4.9G   &  {82.2} 
\\
Swin-T~\cite{liu2021swin} & 224 & 29M  & 4.5G  & 81.3 &
BOAT-Swin-T (ours) & 224 &  31M & 5.2G   &  {82.3}   \\
CSWin-T~\cite{dong2022cswin}  & 224 & 23M  & 4.3G  & 82.7 &
BOAT-CSWin-T (ours)  & 224 & 27M  & 5.1G & \textbf{83.7}   \\
\midrule
ReGNetY-8G~\cite{radosavovic2020designing} & 224 & 39M & 8.0 & 81.7 &
PVTv2-B4~\cite{wang2021pvtv2} & 224 & 62M & 10.1G & 83.6 \\
Twins-B & 224 & 56M & 8.3G & 83.2 &
Shuffle-S~\cite{huang2021shuffle} & 224 & 50M & 8.9G & 83.5 \\ 
NesT-S~\cite{zhang2022nested}, & 224 & 38M & 10.4G & 83.3 &
Focal-S~\cite{yang2021focal} & 224 &  51M & 9.1G   &  {83.5}
\\
Swin-S~\cite{liu2021swin}  & 224 & 50M  & 8.7G  & 83.0  &
BOAT-Swin-S (ours) & 224 & 56M & 10.1G &   {83.6}      \\
CSWin-S~\cite{dong2022cswin}   & 224 & 35M  & 6.9G & 83.6  &
BOAT-CSWin-S (ours)  & 224 &   41M  & 8.0G & \textbf{84.1}   \\
\midrule
ReGNetY-16G~\cite{radosavovic2020designing} & 224 & 84M & 16.0G & 82.9 & 
ViT-B/16T~\cite{dosovitskiy2021image} & 384 & 86M  & 55.4G  & 77.9   \\
DeiT-B~\cite{touvron2020training}  & 224 & 86M  & 17.5G  & 81.8 &
T2T-24~\cite{yuan2021tokens} & 224 & 64M & 14.1G & 82.3 \\
TNT-B~\cite{han2021transformer} & 224 & 66M & 14.1G & 82.8 &
PiT-B~\cite{heo2021rethinking} & 224 & 74M & 12.5G & 82.0 \\
PVTv2-B5~\cite{wang2021pvtv2} & 224 & 82M & 11.8G & 83.8 &
Twins-L & 224 & 99M & 14.8G & 83.7 \\
Shuffle-B~\cite{huang2021shuffle} & 224 & 88M & 15.4G & 84.0 & 
NesT-B~\cite{zhang2022nested}, & 224 & 68M & 17.9G & 83.8 \\
Focal-B~\cite{yang2021focal} & 224 &  90M & 16.0G   &  {83.8}  &
CrossFormer-L~\cite{wang2021crossformer} & 224 &  92M & 16.1G   &  {84.0} \\
Swin-B~\cite{liu2021swin} & 224 & 88M  & 15.4G & 83.5  &
BOAT-Swin-B (ours) & 224 & 98M & 17.8G  & {83.8}      \\
CSWin-B~\cite{dong2022cswin}  & 224 & 78M  & 15.0G  & 84.2 &
BOAT-CSWin-B (ours)  & 224 &   90M  & 17.5G  & \textbf{84.7}   \\
\bottomrule[1pt]
		\end{tabular}
		}
	\end{center}
	\vspace{-3mm}
	\caption{Comparison of image classification performance on the ImageNet-1K dataset.}
	\label{tab:class}
\end{table}

We follow the same training strategies as other vision Transformers. We train our models using the training split of ImageNet-1K~\cite{deng2009imagenet} with $224 \times 224$ input resolution and without external data.  Specifically, both Swin and BOAT-Swin are trained for $300$ epochs, and both CSWin and BOAT-CSWin are trained for $310$ epochs. 
Table~\ref{tab:class} compares the performance of the proposed BOAT models with the state-of-the-art vision backbones. As shown in the table, with a slight increase in the number of parameters and FLOPs,  our BOAT-Swin model consistently improves the vanilla Swin model under the tiny, small and base settings. 
Meanwhile, our BOAT-CSWin model also improves the vanilla CSWin model by a similar degree under the tiny, small and base settings. Such improvements over Swin and CSWin models demonstrate the effectiveness of feature-space local attention. 

\begin{figure*}[h]
    \centering
    \includegraphics[scale=0.5]{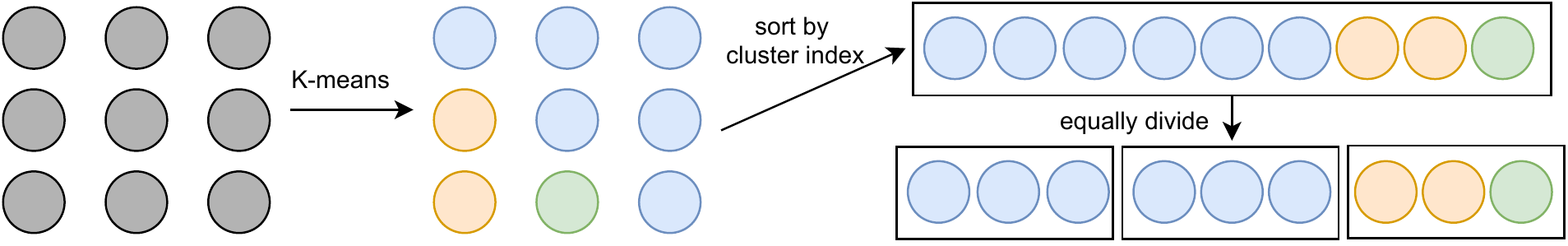}
    \caption{The process of enforcing the clusters from K-means to be equally sized.}
    \label{fig:k-means}
\end{figure*}

\vspace{0.05in}
\noindent\textbf{Comparisons with Reformer and K-means.}
Reformer~\cite{kitaev2020reformer} also exploits feature-space local attention. It divides tokens into multiple groups using Locality Sensitivity Hashing (LSH). based on sign random projections~\cite{charikar2002similarity,li2019sign}, which is independent of specific input data, and might be sub-optimal for different input data. K-means clustering is another choice for dividing tokens into multiple groups for exploiting feature-space local attention. Nevertheless, K-means clustering cannot ensure that the generated clusters are equally sized, which makes it difficult to have efficient parallel implementation on GPU platforms.
To enforce the clusters from K-means clustering to be equally sized, we can sort the tokens according to their cluster index and then equally divide the sorted tokens into multiple groups, as visualized in Figure~\ref{fig:k-means}. However, this sort-and-divide process might divide tokens from a large cluster into multiple groups and also merge tokens from small clusters into the same group. This would negatively impact the overall effectiveness of feature-space local attention.


\begin{table}[!h]
\centering
\begin{tabular}{c|c|c|c}
\hline
Method &  Reformer & K-means &  Ours\\ \hline
Top-1 Accuracy &    $81.7$    &  $81.8$     &      $82.3$           \\ \hline
\end{tabular}
\caption{Comparison of image classification accuracy with Reformer and K-means.}
\label{ref}\vspace{-0.1in}
\end{table}

We compare the performance of BOAT-Swin-Tiny against the performance of a model where our balanced hierarchical clustering is replaced with LSH in Reformer or K-means clustering. We keep the other layers unchanged. As shown in Table~\ref{ref}, our BOAT-Swin-Tiny clearly outperforms Reformer and K-means clustering.

\vspace{0.05in}
\noindent\textbf{The effectiveness of FSLA.} To directly demonstrate the effectiveness of feature-space local attention (FSLA), we replace all FSLA blocks in BOAT-Swin-T with image-space local attention (ISLA) blocks. As shown in Table~\ref{replace}, the accuracy drops from $82.3\%$ to $81.5\%$.

\begin{table}[h!]
\centering
\begin{tabular}{c|c|c}
\hline
Model &  BOAT-Swin-T (with FSLA) & Baseline (with ISLA) \\ \hline
Accuracy & $82.3$               &   $81.5$                      \\ \hline
\end{tabular}
\caption{Ablation study on FSLA by replacing FSLA with ISLA.\vspace{-3mm}}
\label{replace}\vspace{-0.05in}
\end{table}

\vspace{0.05in}
\noindent\textbf{The effectiveness of overlapping balanced hierarchical clustering.} 
We compare the performance of overlapping balanced hierarchical clustering with its non-overlapping counterpart on the ImageNet-1K dataset. As shown in Table~\ref{overlap}, the overlapping setting achieves consistently higher classification accuracy in BOAT-CSwin-Tiny, Small and Base models. Higher accuracy is expected since the overlapping setting gives rise to larger receptive fields.

\begin{table}[!h]
\centering
\begin{tabular}{c|c|c|c}
\hline
Overlap & BOAT-CSWin-T & BOAT-CSWin-S & BOAT-CSWin-B \\ \hline
No      &  $83.3\%$               &   $84.0\%$               &      $84.5\%$           \\ 
Yes    &     $83.7\%$            &    $84.1\%$               &       $84.7\%$          \\ \hline
\end{tabular}
\caption{Comparison of image classification accuracy between overlapping balanced hierarchical clustering and the non-overlapping version.\vspace{-3mm}}
\label{overlap}\vspace{-0.05in}
\end{table}

\vspace{-0.05in}
\subsection{Semantic Segmentation}
\vspace{-0.05in}

We further investigate the effectiveness of our BOAT for semantic segmentation on the ADE20K dataset~\cite{zhou2017scene}. Here, we employ UperNet~\cite{xiao2018unified} as the basic framework.
For a fair comparison, we follow previous work and train UperNet 160K iterations with batch size $16$ using $8$ GPUs. In Table~\ref{tab:seg}, we compare the semantic segmentation performance  of our BOAT with other vision Transformer models including  Swin~\cite{liu2021swin}, Twins~\cite{chu2021twins}, Shuffle Transformer~\cite{huang2021shuffle},  Focal Transformer~\cite{yang2021focal},  and CSWin~\cite{dong2022cswin}. As shown in the table, with a slight increase in the number of parameters and FLOPs, our BOAT-Swin model consistently improves the semantic segmentation performance of the Swin model under the tiny, small and base  settings. 
Meantime, our BOAT-CSWin also constantly obtains higher segmentation mIoUs than the CSWin model under the tiny, small and base settings.

\begin{table}[!t]\small
	\setlength{\abovecaptionskip}{0pt}
	\setlength{\belowcaptionskip}{0pt}
\begin{center}
\resizebox{0.98\linewidth}{!}{
\begin{tabular}{@{}lccc|lccc@{}}
\toprule[1pt]
Method  &  \#para.(M)  & FLOPs(G)  & mIoU($\%$) & Method  &  \#para.(M)  & FLOPs(G)  & mIoU($\%$)  \\ 
\midrule
TwinsP-S~\cite{chu2021twins} & 55  & 919  & 46.2  &
Twins-S~\cite{chu2021twins} & 54  & 901  & 46.2   \\
Shuffle-T~\cite{huang2021shuffle} & 60 & 949 & 46.6 &
Focal-T~\cite{yang2021focal} & 62 & 998 & 45.8 \\
Swin-T~\cite{liu2021swin} & 60  & 945  & 44.5  &
BOAT-Swin-T~(ours)& 62  & 986  & 46.0   \\
CSWin-T~\cite{dong2022cswin} & 60  & 959  & 49.3   &
BOAT-CSWin-T~(ours) & 64  & 1012  & \textbf{50.5}   \\

\midrule
TwinsP-B~\cite{chu2021twins} & 74  & 977  & 47.1  &
Twins-B~\cite{chu2021twins} & 89 & 1020  & 47.7   \\
Shuffle-S~\cite{huang2021shuffle} & 81 & 1044 & 48.4 &
Focal-S~\cite{yang2021focal} & 85 & 1130 & 48.0 \\
Swin-S~\cite{liu2021swin} & 81  & 1038  & 47.6  &
BOAT-Swin-S~(ours)& 87  & 1113  & 48.4   \\
CSWin-S~\cite{dong2022cswin} & 65  & 1027  & 50.0  &
BOAT-CSWin-S~(ours) & 70 & 1101 & \textbf{50.6}  \\

\midrule
TwinsP-L~\cite{chu2021twins} & 92  & 1041  & 48.6   &
Twins-L~\cite{chu2021twins} & 133  & 1164 & 48.8   \\
Shuffle-B~\cite{huang2021shuffle} & 121 & 1196 & 49.0 &
Focal-B~\cite{yang2021focal} & 126 & 1354 & 49.0 \\
Swin-B~\cite{liu2021swin} & 121  & 1188  & 48.1  &
BOAT-Swin-B~(ours)&  131 &  1299 & 48.7   \\
CSWin-B~\cite{dong2022cswin} & 109   & 1222 &  50.8 &
BOAT-CSWin-B~(ours) & 121  & 1349 & \textbf{50.9}   \\

\bottomrule[1pt]
		\end{tabular}}
	\end{center}
	\vspace{-3mm}
	\caption{Performance of semantic segmentation on ADE20K.  FLOPs are obtained at $512\times 2048$ resolution. mIoU is for the single-scale setting. Testing image size is $512\times 512$.}
	\vspace{-0mm}
	\label{tab:seg}
\end{table}

\subsection{Object Detection} 
We also evaluate the proposed BOAT on object detection. Experiments are conducted on the MS-COCO dataset using the Mask R-CNN~\cite{he2017mask} framework.  Since CSWin has not released codes for object detection, we only implement BOAT-Swin for this task.  We adopt the $3\times$ learning rate schedule, which is the same as Swin. We compare the performance of our BOAT-Swin and the original Swin in Table~\ref{tab:det}. The evaluation is on the MSCOCO val2017 split. Since Swin only reports the performance of  Swin-Tiny and Swin-Small models when using the Mask R-CNN framework, we also report the performance of our BOAT-Swin-Tiny and  BOAT-Swin-Small only. As shown in Table~\ref{tab:det}, with a slight increase in the number of parameters and FLOPs, our BOAT-Swin consistently outperforms the original Swin. 
\begin{table}[h]\small
	\setlength{\abovecaptionskip}{0pt}
	\setlength{\belowcaptionskip}{0pt}
\begin{center}
\begin{tabular}{@{}lcccccc@{}}
\toprule[1pt]
Method  &  \#para.(M)  & FLOPs(G)  & mAP$^{\mathrm{Box}}$  &  mAP$^{\mathrm{Mask}}$ \\ 
\midrule

Swin-T & 48  & 267  & 46.0 & 41.6  \\
BOAT-Swin-T~(ours)&  50 & 306  & \textbf{47.5} & \textbf{42.8}   \\\midrule
Swin-S & 69  & 359  & 48.5 & 43.3   \\
BOAT-Swin-S~(ours)& 75  & 431 & \textbf{49.0} & \textbf{43.8}  \\

\bottomrule[1pt]
		\end{tabular}
	\end{center}
	\vspace{-3mm}
	\caption{Performance of object detection on the MS-COCO dataset.  FLOPs are obtained at $800 \times  1280$ resolution.}
	\label{tab:det}
	\vspace{-0.1in}
\end{table}

\section{Conclusion}
In this paper, we have presented a new Vision Transformer architecture named Bilateral lOcal Attention Transformer (BOAT), which performs multi-head local self-attention in both feature and image spaces. 
To compute feature-space local attention, we propose a hierarchical balanced clustering approach to group patches into multiple evenly sized clusters, and self-attention is computed within each cluster.
We have applied BOAT to multiple computer vision tasks including image classification, semantic segmentation and object detection. Our systematic experiments on several benchmark datasets have demonstrated that BOAT can clearly and consistently improve the performance of existing image-space local attention vision Transformers, including Swin~\cite{liu2021swin} and CSWin~\cite{dong2022cswin}, on these tasks.

\bibliography{refs_scholar,egbib}
\end{document}